\documentclass{ieeeaccess}
\usepackage{booktabs}
\usepackage{cite}
\usepackage{amsmath,amssymb,amsfonts}
\usepackage{algorithmic}
\usepackage{graphicx}
\usepackage{textcomp}
\def\BibTeX{{\rm B\kern-.05em{\sc i\kern-.025em b}\kern-.08em  T\kern-.1667em\lower.7ex\hbox{E}\kern-.125emX}}
\usepackage{hyperref}
\usepackage{multirow}
\usepackage[]{footmisc}
\graphicspath{{.}{./Figure/}}

\begin{document}
\history{}
\doi{}
\title{A Review of the Trends and Challenges in Adopting Natural Language Processing Methods for Education Feedback Analysis}

\author{%
        \IEEEauthorblockN{
            Thanveer Shaik \thanks{T.~Shaik, Thanveer.Shaik@usq.edu.au,  orcid=0000-0002-9730-665X}%
            \IEEEauthorrefmark{1},
            Xiaohui Tao \thanks{X.~Tao, Xiaohui.Tao@usq.edu.au, orcid=0000-0002-0020-077X}%
            \IEEEauthorrefmark{1},
            Yan Li \thanks{Y.~Li, 	Yan.Li@usq.edu.au, orcid=0000-0002-4694-4926}
            \IEEEauthorrefmark{1},
            Christopher Dann \thanks{C.~Dann, Chris.Dann@usq.edu.au, orcid=0000-0001-7477-0305}%
            \IEEEauthorrefmark{2},
            Jacquie McDonald \thanks{J.~McDonald, Jacquie.McDonald@usq.edu.au}%
            \IEEEauthorrefmark{3},
            Petrea Redmond \thanks{P.~Redmond, Petrea.Redmond@usq.edu.au, orcid=0000-0001-9674-1206}%
            \IEEEauthorrefmark{2},
            Linda Galligan\thanks{L.~Galligan, 	Linda.Galligan@usq.edu.au, orcid=0000-0001-8156-8690}%
            \IEEEauthorrefmark{1}
        }
        \\
        \IEEEauthorblockA{
            \IEEEauthorrefmark{1}
            School of Mathematics, Physics and Computing, University of Southern Queensland, Toowoomba
            \\
            \IEEEauthorrefmark{2}
            School of Education, University of Southern Queensland, Toowoomba
            \\
            \IEEEauthorrefmark{3}Academic Development, University of Southern Queensland, Toowoomba
            }
        }

\markboth
{Shaik \headeretal: Preparation of Papers for IEEE TRANSACTIONS and JOURNALS}
{Shaik \headeretal: Preparation of Papers for IEEE TRANSACTIONS and JOURNALS}

\corresp{Corresponding author: Thanveer Shaik (e-mail: Thanveer.Shaik@usq.edu.au).}

\begin{abstract}
Artificial Intelligence (AI) is a fast-growing area of study that stretching its presence to many business and research domains. Machine learning, deep learning, and natural language processing (NLP) are subsets of AI to tackle different areas of data processing and modelling. This review article presents an overview of AI’s impact on education outlining with current opportunities. In the education domain, student feedback data is crucial to uncover the merits and demerits of existing services provided to students. AI can assist in identifying the areas of improvement in educational infrastructure, learning management systems, teaching practices and study environment. NLP techniques play a vital role in analyzing student feedback in textual format. This research focuses on existing NLP methodologies and applications that could be adapted to educational domain applications like sentiment annotations, entity annotations, text summarization, and topic modelling. Trends and challenges in adopting NLP in education were reviewed and explored. Context-based challenges in NLP like sarcasm, domain-specific language, ambiguity, and aspect-based sentiment analysis are explained with existing methodologies to overcome them. Research community approaches to extract the semantic meaning of emoticons and special characters in feedback which conveys user opinion and challenges in adopting NLP in education are explored. 


\end{abstract}


\maketitle

\section{Introduction}\label{sec:Introduction}
Artificial Intelligence (AI) is a fast-growing topic with its cognitive human-like intelligence in building decision-making systems. AI can revolutionize education with its capacity for prediction and classification by processing huge amounts of structured data sets such as SQL databases and unstructured datasets such as videos and audios. AI introduces machine learning methodologies to personalize the student learning experience via learning management systems ~\cite{Peters2017}, deep learning, and transfer learning to use pre-trained concepts to deal with new similar problems~\cite{hunt2017transfer}, natural language processing (NLP) methods~\cite{Kastrati2021} to listen to student feedback, process them and output predictive insights on their opinion towards learning infrastructure. AI can transform existing educational infrastructures~\cite{1} namely online tutoring, learning management systems, curriculum, employment transitions, teacher training, assessments, and research training. The institutional project data are diverse and inclusive of student feedback in textual format class-room recordings in video and audio formats.  

Chassignol et al.~\cite{Chassignol2018} defined AI as an “Artificial Intelligence is that activity devoted to making machines intelligent, and intelligence is that quality that enables an entity to function appropriately and with foresight in its environment”. Educational institutions have extensively adopted AI in different forms of service delivery to students ~\cite{9069875}. One of the most widely used AI methodologies for student opinion mining is NLP~\cite{BarrnEstrada2020}. It plays a key role in interpreting feedback or opinions of end-users. Most institutions in the world invest their time and resources to understand end-users' feedback. NLP can read the feedback in most languages without much human intervention and can analyze textual data and unwrap the end-user perception and opinion on a service, product, or human. In recent years, NLP has been applied to review movies, books, gadgets and so on~\cite{Yadav2019}. Topic modelling techniques are part of NLP to read text corpus and can summarize, annotate or categorize text documents. Furthermore, it uses various techniques like part-of-speech (POS) tagging to understand the context of words. 

Eggert\cite{eggert2021artificial} discussed the opportunities of AI in education. The author proposed an AI method to improve teaching methods by collecting vast amounts of data related to each student’s prior knowledge, emotional state, or economic background and adjusting the teaching approach through adaptive learning platforms (ALP). Intelligent tutoring systems (ITS) is one of the ALP components. With automation of repeated tasks would allow teaching staff to design new instructional approach and focus on non-routine work. The other opportunity discussed in that article is to expose students to some AI-driven tools to cope with the future labour world that is highly dependent on technologies and focus on lifelong learning via improved access to Massive Open Online Courses (MOOCs). AI can enhance student’s learning experience in MOOCs by identifying areas where personalized guidance is required. Holstein et al.~\cite{Holstein2019} also stressed the need for personalized guidance to students in their work on AI-enhanced classrooms. Using participatory speed dating (PSD)~\cite{Bhimdiwala2021}, the authors found real-time support was needed from the AI system to identify when a student needs a human’s help for motivation. Holstein et al.~\cite{Holstein2019real} also focused on the challenges of involving non-technical stakeholders due to the complexity of learning analytics systems~\cite{labarthe2018analyzing}. The authors proposed Konscia, a wearable and real-time awareness tool for teachers working in AI-enhanced K-12 classrooms. In addition, they demonstrated the process of non-technical stakeholders’ participation in designing a complex learning analytics system. Alrajhi et al.~\cite{alrajhi2021urgency} stressed the need to analyse student feedback or comments in MOOC as it would help to understand the student need for intervention from instructors.

Chen et al.~\cite{9069875} surveyed the impact of AI on education. The authors discussed the technical aspects of AI in education: assessment of students and schools, grading and evaluating papers and exams, smart schools, personalized intelligent teaching, online and mobile remote education. Their study scope was confined to the application and effects of AI in administration, teaching, and learning. To enable instructors and teachers with effective grading capabilities, an adaptive learning method was used in applications of Knewton and ensured a continuous student improvement in learning~\cite{sharma2019landscape}. Applications like Grammarly, Ecree, Paper-Rater and Turnitin leverage AI to assist educational institutions and teachers in performing plagiarism checks, typographical and grammatical error checks. The student learning experience is an essential aspect of the education domain. AI enables an adaptive learning system for students based on their backgrounds to assist in tracking their learning progression and enhance the system to customize the content according to student’s needs to leverage a personalized system. A quick interactive system using AI would reduce the gap between students and educational providers and assist in listening to students’ opinions and queries. 

With the extensive research being conducted in analyzing AI's impact on education~\cite{gulson2018education, pedro2019artificial} and discovering the opportunities in the education domain, educational institutions have focused on building a cognitive intelligent system using AI. In this process, the foremost step is to listen to students’ opinions and feedback on existing educational infrastructure, teaching practices, and learning environments. In academic institutions, it is traditional practice to request student feedback to gather students' perception of the teaching team and their learning experience in the course. The student feedback could be in quantitative or qualitative formats, using numerical answers to rate the performance or textual comments to questions~\cite{8190703}. Monitoring and tracking students’ feedback manually is a time-consuming and resource-demanding task. NLP can contribute to this task with its annotation and summarization capabilities. This study reviewed NLP methodologies that can contribute to the education domain, and the following research questions were explored:

\begin{itemize}
    \item What are the existing methodologies being used for NLP?
    \item What are the generic challenges of using NLP in the education domain?
    \item What are the current trends of NLP in student feedback analysis?
    \item How can NLP methodology in other disciplines be adopted to the education domain? 
\end{itemize}

Machine learning and deep learning are part of AI methodologies. Machine learning is a set of algorithms can analyze data, learn and apply. Deep learning techniques  holds multi-layer neural network with processing layers to train new concepts and link to previously known concepts. Deep learning enhances NLP with concepts like continuous-bag-of-words and skip-gram model. Convolutional neural networks (CNNs) ~\cite{Li2018},  recurrent neural networks (RNNs), their special cases of long short-term memory (LSTM) and gated recurrent units (GRUs) are different forms of deep learning techniques used in text classification~\cite{ramaswamy2018customer}~\cite{Prokhorov2019}. In this article, existing works using the AI methodologies to analyze text data are explored. Although few research works are not directly related to student feedback, the methods can be adopted to students' feedback analysis.

The contributions of this research are as follows:
\begin{itemize}
    \item Enhanced understanding of the impact of AI on education with open opportunities in the industry. 
    \item Synthesis of existing NLP methodologies to student user feedback and annotate their views. 
    \item Exploring trends and challenges in NLP that need to be addressed to be adopted to the education domain. 
\end{itemize}

The remainder of the paper is organized into sections. Section~\ref{Methods} defined feature extraction, feature selection, and topic modelling techniques with other researchers’ work. Text evaluation techniques like summarization, knowledge graphs, annotation, existing NLP methodologies being used for NLP are defined. In Section~\ref{challenges}, challenges in adopting NLP in the education domain are discussed. Section~\ref{Discussion}  presents a discussion on this work. The article concludes with limitations and future work of the study presented in Section~\ref{Conclusions}

\section{Methodology} \label{Methods}
Feature extraction and feature selections are mandatory  data preprocessing steps to transform text data into quantitative vector formats before feeding the students' feedback data to traditional machine learning algorithms or machine learning techniques like topic modelling. In this section, existing methods in feature extraction, feature selection, and topic modelling will be discussed.

\subsection{Feature Extraction}\label{featureextraction}
Feature extraction techniques can be applied to prepare the students' feedback data and transform it for machine learning modelling. For example, in NLP, there are feature extraction techniques like Bag of Words (BoW), Term Frequency (TF)-Inverse Document Frequency (IDF), and Word Embedding~\cite{Ahuja2019}. 

\textbf{Bag of Words (BoW)}~\cite{Balamurali2020} is a common feature extraction method that involves a vocabulary of known words and a measure of the presence of known words. The BoW is only concerned with known words in a document. It will not consider the structure or order of words in a document that ignores the context of the words~\cite{goldberg69neural}. TF-IDF~\cite{ArroyoFernndez2019} estimates the importance of each word or term in a document based on their weights~\cite{Shah2020}. IDF of a word gives how common or rare a word is in a corpus. The closer the value is to zero, the more common a word is in a corpus. TF-IDF is a multiplication of TF and IDF. 

\textbf{Word Embedding}~\cite{Onan2020} is a learned representation of text with similar meaning. It enhances the generalization process and reduces dimensionality. The most common word embedding techniques are Word2Vec, GloVe, Doc2Vec~\cite{Amin2020} and Bidirectional encoder representations from transformers (BERT)~\cite{devlin2018bert}. 
Word2vec algorithm is built on a neural networks model to learn word associations from a large corpus of text. The trained algorithm can detect synonymous words or even suggest additional words for a partial sentence. Word2Vec generates the number of dimensions for each word in a corpus and then searches at the context level of the occurrence of the words in a sentence. In a vector space, all the words with similar contexts are grouped. A GloVe approach combines the matrix factorization technique and latent semantic analysis (LSA) with a context-based learning in Word2Vec. Doc2Vec is a tool to create vector or numeric representations of documents. BERT is a pre-trained deep bidirectional representations from unlabeled text by jointly conditioning on both left and right context in all layers. BERT can perform word or sentence embedding to extract vectors from text data. It has an advantage over techniques like Word2Vec where each word has a fixed representation in Word2Vec irrespective of context. BERT can produce word representations dynamically based on words around them~\cite{hou2020bert}.

Waykole et al.~\cite{waykole2018review} evaluated text classification based on feature extraction techniques such as bag of words, TF-IDF, and Word2Vec. In that study, each of the feature extraction techniques were evaluated with machine learning algorithms, for example logistic regression and random forest classifier with 3-fold stratified cross-validation. The experimentation results showed that Word2Vec was a better feature extraction technique with a random forest classifier was better for the text classification. Similarly, count vectorizer, TF-IDF and Word2Vec techniques were compared using a logistic regression model. Deepa et al.~\cite{Deepa2019} proposed an approach to detect the polarity of words from Twitter using three feature extraction techniques count vectorizer, Word2Vec, TF-IDF and two dictionary-based methods of valence aware dictionary and sentiment reasoner (VADER) and SentiWordNet. Feature extraction techniques achieved better accuracy than dictionary-based methods. For example, count vectorizer achieved the highest classification accuracy of 81\%. Twitter text is short text analysis which is similar to students' feedback to an open-ended question in an educational institution where feature extraction techniques can be adopted.

TF-IDF feature extraction generates feature vectors with high dimensions in a large text corpus~\cite{Dzisevic2019}. In the study~\cite{Dzisevic2019}, the TF-IDF extraction technique was evaluated by adding dimensionality reduction techniques, latent semantic analysis (LSA) and linear discriminant analysis (LDA). Using a neural network classifier, the authors compared the classification performance of plain TF-IDF, TF-IDF LSA, and TF- IDF LDA methods on short texts. The research outcome stated that the TF-IDF approach outperformed the other two approaches with larger datasets. With smaller datasets, the TF-IDF and TF-IDF LSA achieved similar accuracy. However, the TF-IDF LDA approach had difficulty accurately classifying the text, as it failed to reduce the noise. 

Deep learning techniques were used to evaluate word embedding techniques of Word2Vec and GloVe. In a study by~\cite{Goularas2019}, CNNs and RNNs were compared, and ensembles a combination of CNN and LSTM networks and compared. Eight different combinations of deep learning algorithms were implemented. Their comparison results showed that the GloVe system enhanced the performance by about 5-7\% compared to Word2Vec. Sangeetha et al.~\cite{Sangeetha2020} proposed a novel approach to analyze and find students' emotions in their feedback. In the study, feedback sentences were processed parallel across a multi-head attention layer with embedding techniques GloVe and contextualized vectors (Cove). The proposed method was tested with dropout rates to improve the accuracy. The authors compared the performance of proposed methods' with baseline models LSTM, LSTM+ATT, multi-head ATT, and fusion in terms of accuracy 86.27\%, 87.49\%, 90.03\%, and 94.13\% respectively.


Zhang et al.~\cite{zhang2020bert} proposed a fine-tuned BERT model for sentiment analysis of student feedback to courses. In that study, intra domain unsupervised training was performed using the BERT model. To add grammatical constraints to the output of BERT model, a conditional random field layer was introduced. In addition, binding corporate rules — double attention layers were added to target sentiment analysis of the student feedback. Masala et al.~\cite{masala2021extracting} analyzed student feedback provided to each course and extract important ideas on various components. The authors used the BERT model to extract keywords from student feedback from each course, find contexts for repeated keywords, and group similar contexts. With this approach, 59\% of the feedback text was reduced at a cost if mean average error increased to 0.06 while predicting course ratings from student feedback. Wu et al.~\cite{wu2021word} proposed pre-trained word embeddings to automatically create clusters such as homogeneous and heterogeneous student groups based on students' knowledge. Homogeneous groups can assist teachers to provide collective feedback, and heterogeneous groups can support and improve collaborative learning.

Feature extraction methods normally break down students' feedback data into word tokens to prepare the data for semantic and grammatical analysis. Neural network-based BoW, TF-IDF gives the frequency of words in a document, word embedding techniques like Word2Vec, GloVe, Cove, and BERT reduce the dimensionality of a word to group similar contexts. The performances of the feature extraction methods are often compared using machine learning and deep learning methods in~\cite{Onan2020, Amin2020,waykole2018review,Deepa2019,Dzisevic2019,Goularas2019,8999624,zhang2020bert,masala2021extracting,wu2021word}. 


\subsection{Feature Selection}
Feature selection is a process of reducing data dimensionality in terms of features. This would maintain or enhance the performance of a machine learning algorithm. The reduction criteria would simplify a model’s complexity and consistently maintain accuracy. Considering n features in a dataset, the number of the feature subsets would be $2^n$. An increase in the features count would make the modelling infeasible~\cite{ghojogh2019feature,cai2018}. The stability or robustness of feature subsets was evaluated by grouping similar features or considering all feature subsets, removing the non-contributing features, and the size of the feature subsets. The feature subset evaluation methods are broadly categorized as filters, wrappers, or embedded methods~\cite{kou2020}~\cite{Nikoli2020}.

\begin{figure}[!htp]
    \centering
    \includegraphics[width=\columnwidth]{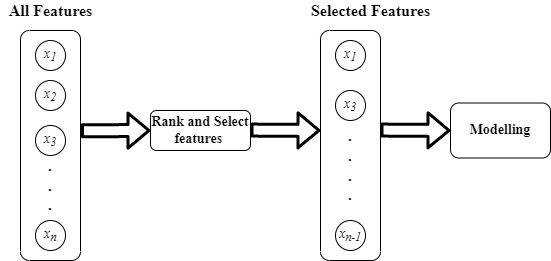}
    \caption{Filter Methods}
    \label{fig:filtermethods}
\end{figure}
\subsubsection{Filter Methods}
Filter methods rank the key features and select high representative features by setting a threshold~\cite{bommert2020}. As shown in Figure~\ref{fig:filtermethods}, filter methods rank the features and select them before actual modelling. In addition, this technique filters the low importance features before training a model. The feature importance technique assesses two measures in ranking the features. The first measure is to check the predictive power of each feature toward the target variable(s). These are called correlation criteria or dependence measures. Mutual information, ${\chi}^2$ statistic, Markov blank, and minimal-redundancy-maximal-relevancy techniques extract a feature's correlation with a target variable. The second measure in the feature importance technique is redundancy, which assesses the features with redundant information. This detects the redundant features by evaluating relevant measures among the independent variables. An article by Wang~\cite{wang2020} presented a redundant feature analysis. Its process is to find the most relevant features in predicting the target variables and use the relevant features to estimate the redundancy in other features.

\begin{figure}[!htp]
    \centering
    \includegraphics[width=\columnwidth]{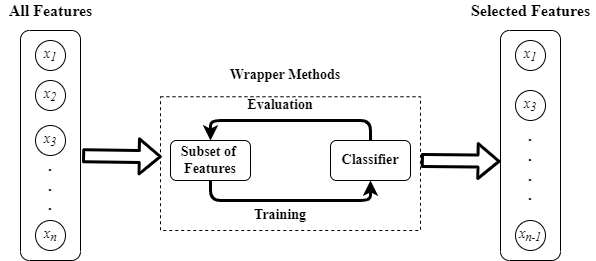}
    \caption{Wrapper Methods}
    \label{fig:wrappermethods}
\end{figure}
\subsubsection{Wrapper Methods}
Wrapper methods search for a subset of features using a predefined classifier and then the performance of the subset of features is evaluated using predefined classifiers~\cite{bommert2020}. In wrapper methods, a machine learning algorithm is used to enhance the feature selection performance. As shown in Figure~\ref{fig:wrappermethods}, a subset of features is selected and trained by a classifier with the selected features. Then, the performance of the classifier is evaluated. Sequential forward selection (SFS) is an example of a wrapper method with sequential feature selection methods. It is a greedy search algorithm that extracts an optimal subset of features iteratively based on the classifier performance. Features are selected one-by-one from the pool of all features iteratively.

\begin{figure}[!htp]
    \centering
    \includegraphics[width=\columnwidth]{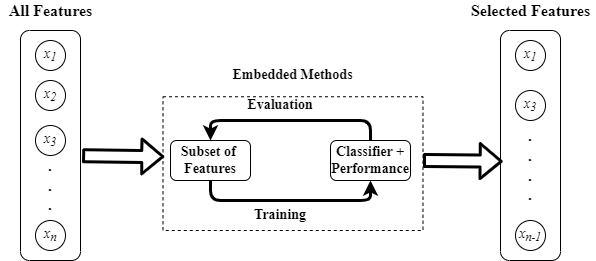}
    \caption{Embedded Methods}
    \label{fig:embeddedmethods}
\end{figure}

\subsubsection{Embedded Methods}
Embedded methods normally combine a filter method and a wrapper method~\cite{dhal2021}. They overcome the challenges of low accuracy in filter methods and slow computation speed in wrapper methods. Embedded methods analyze the optimum features, contributing to the classifier’s accuracy. As shown in Figure~\ref{fig:embeddedmethods}, embedded methods estimate the performance of each subset of features. One of the most common embedded methods is regularization, which is to reduce the degree of overfitting or variance of a model by adding a penalty against its complexity for L1 regularization methods~\cite{osman2017}. 

Parlar et al.~\cite{parlar2018qer} proposed a query expansion ranking (QER) method for feature selection. They compared it with other feature selection methods like information gain, Chi-square, document frequency difference, and optimal orthogonal centroid using classifiers like Naïve Bayes-multinomial, a support vector machine, maximum entropy modelling, and a decision tree. The authors tested the model performances on English and Turkish review databases. The proposed feature selection method outperformed the other feature selection methods in the Naïve Bayes multinomial classifier. Similar techniques were adopted to analyze student feedback in teaching evaluation system by Pong-Inwong et al.~\cite{7924899}. In the study, filter method was opted for feature selection and number of attributes in the data were reduced to 18 based on Chi-Square value. Three machine learning algorithms ID3, J48 and Naïve Bayes were used for student feedback classification and compared their performance with vote ensemble learning. The voting ensemble learning integrated with Chi-Square feature selection outperformed traditional machine learning algorithms with an accuracy of 87.16\%.

Gutiérrez et al.~\cite{gutierrez2018mining} proposed social mining model architecture to increase the quality of learning and e-learning based on students' feedback analysis. This approach was focused to enhance teaching techniques and recommend courses for teacher improvement in higher education. As part of feature selection process in the study, random forest importance measure method was used with which weights of each word can be computed and filter them based on higher weights. The selected features were passed to SVM with kernels linear, radial, poly, and random forest classifiers. The machine learning classifiers were trained using k-fold cross validation and SVM model with radial kernel outperformed other models with an accuracy of 85.17\%. Similarly, Soukaina et al.~\cite{9268731} proposed an information gain filter method to select most relevant features in students' feedback in an optimized sentiment analysis approach. SVM, random forest, and Naive Bayes classifiers were used in the study and compared their performances before feature selection and after selection. Random forest dominated the other two classifiers before the selection with an accuracy of 81.6\% and SVM outperformed with an accuracy of 85.9\% after the feature selection. 



Feature selection approaches are better with noise resistance and can help to avoid noise or irrelevant data for data modelling. Other modern feature selection methods were proposed and compared with existing methods using machine learning methods in~\cite{parlar2018qer,8483345}

\subsection{Topic Modelling}
Topic modelling automatically analyzes a corpus of documents with text data techniques using machine learning techniques and determines cluster words~\cite{Wallach2006,Patil2018}. The technique does not need any training to cluster the words from the corpus. This is an unsupervised machine learning technique~\cite{nikolenko2016}. Topic modelling divides a corpus of documents into groups to extract a list of topics covered, and several sets of documents are grouped by the topics they covered. The topic modelling techniques are broadly categorized into probabilistic and non-probabilistic models~\cite{Curiskis2020,Asmussen2019}.

\subsubsection{Non-probabilistic Models}
Non-probabilistic models are matrix factorization algebraic approaches. These models came into use with latent semantic analysis (LSA) and Non-negative matrix factorization (NMF)~\cite{8250563}. Both LSA and NMF mechanisms work on BoW approaches. As discussed in Section~\ref{featureextraction}, BoW converts a corpus into a term-document matrix to extract the frequency of the terms and ignores the order of the terms. LSA is an algebraic method that generates a matrix with words presented in a corpus. It assumes that words that are similar in meaning will occur very close in the text~\cite{Landauer1998}. The technique is based on single value decomposition (SVD) which reduces the number of words while preserving a similar structure. The similarity of the texts will be computed using vector representation and organized into semantic clusters.NMF transforms high dimensional data into low dimensional data with no negative components and clusters simultaneously~\cite{Lee1999}. This is also called positive matrix factorization (PMF). It is an unsupervised machine learning technique that can extract relevant information without previous insights into the original data.

\subsubsection{Probabilistic Models}
Probabilistic models are fully unsupervised approaches that are tweaked to guide in latent dirichlet allocation (LDA) modelling and semi-supervised learning in a probabilistic latent semantic analysis ~\cite{chen1996building}. Probabilistic latent semantic analysis (PLSA) is to detect semantic co-occurrence of words or terms in a corpus~\cite{Hofmann2001}. This is built based on the first statistical model, a model that revealed the semantic co-occurrence in a document term matrix of the corpus. Due to its unsupervised nature, PLSA is capable of determining the number of topics, the probability of a topic and the probability of a document containing the topic. It groups unknown topics of every existing document. LDA is a commonly used technique in topic modelling which is built based on De Finetti’s theorem, which states that positively correlated exchangeable observations are conditionally independent relative to some latent variable~\cite{de2017theory}. It can capture inter and intra document statistical structures on assumptions that a corpus has a predefined number of topics and each document in the corpus has a different proportion of the topics. It is a hidden variable model which uncovers hidden patterns in gathered data in a corpus.

An LDA technique based on a topic modelling methodology was selected in mobile learning research to find the topic trends~\cite{hamzah2020discovering}. Out of 50 topics extracted from the LDA, 25 topics were selected and grouped into three dimensions of technology, learning and learners in that mobile learning. Similarly, as part of designing a course structure for virtual reality with augmented reality and mixed or extended reality, the LDA technique was employed in the research study~\cite{onah2021mooc}. The study was to understand the motive of learners (students) in joining the course using topic modelling. It revealed that learners had little experience in designing virtual applications. Also, the learners had little experience in a programming language. Designing a massive open online course without understanding learners’ engagement would lead to a high dropout rate.  

To enhance computer science course teaching materials, Marcal et al.~\cite{marccal2020strategy} proposed an innovative approach for extracting topics from StackOverflow, a question-and-answer website for professional and enthusiastic programmers, to identify unknown or misunderstood topics. Using these topics, to transform the course teaching material, the authors classified the question types into eight categories: debugging, how to, what, is there, possible, looking, advise, and optimal using an SVM. The LDA technique for topic modelling generates five topics for each of the eight types of questions. Based on the keywords in all five topics in each question type, professors or lecturers could compare and enhance their material to fill the gap. Course satisfaction surveys were analyzed to extract student opinions by using the LDA technique~\cite{Unankard2020}. In a study by Cunningham et al.~\cite{cunningham2019visualizing}, nine different topics or aspects were selected using a topic modelling technique. Each student comment was separated into ideas to calculate sentiment and also overall sentiment and satisfaction was considered. The authors visualized one-course feedback over different semesters in terms of aspects like tutorial, lecture, assignment, content, and lecturer. 

To analyze international students' needs and perceptions, and grouping them into categories~\cite{Singh2020}, Adriana et al.~\cite{perez2018international} proposed a probabilistic topic model approach using LDA. The authors used a machine learning for language toolkit (MALLET)~\cite{Liermann2021} to run LDA and selected 20 topics based on 59,662 reviews. The topics covered in the research were \textit{language skills, convenient accommodation, weather, academic burdens, interesting courses} and so on. The topics were ordered by their weight in the composition of the whole set of reviews by its importance on what comprises a good university, living expenses, sound teaching, expensive country, and city offerings. As part of the strategic planning of a university to increase student enrollment, knowledge mining on online reviews was performed using ensemble LDA (eLDA) in a study~\cite{srinivas2019topic}. The authors split the database into training data and held out data where the training data to LDA to extract the probabilistic score of words related to each topic being generated. To avoid inconsistency in the LDA results due to its Collapsed Gibbs Sampling (CGS), multiple LDA models were trained in parallel and stored the results in a database for further sentence labelling. The held-out data were labelled using the trained LDA model. Further, the held-out data were manually annotated with prior knowledge of identified topics in the database. Based on the top five values in each topic, 12 meaningful topics like academic support, diversity, faculty, financial aid, the weather were categorized. 

Pyasi et al.~\cite{8658457} developed a student feedback analysis tool to extract sentiments and suggestions from students' feedback using sentiment analysis models, NLP techniques, LDA, and visualization techniques. In this study, Textblob~\cite{loria2018textblob} and polarity analyzer were used for sentiment analysis and Textblob dominated with Recall is 96.17\%, Precision is 67.47\% and F-score of 79.30\%. Generalized linear models (GLM), SVM, conditional inference tree (CTREE) and decision tree C5.0 classification models were used for suggestion extraction and C5.0 model has better other classifiers with recall is 80.2\%, precision is 77.5\% and F-Score is 78.1\%. For topic modelling, LDA and k-means clustering with cosine similarity scores were compared and LDA model was capable to extract multiple topics on single student comment whereas cosine clusters assigned a single topic to the comment. 

Curiskis et al.~\cite{Curiskis2020}, proposed an evaluation of document clustering and topic modelling methods in online social media networks like Twitter and Reddit~\cite{8681445,Ruan2022}. The authors used four feature representation techniques Doc2vec, weighted Word2vec, unweighted Word2vec, and TF-IDF on three benchmark databases extracted from Twitter and Reddit API. For document clustering, k-means clustering~\cite{Rani2021}, k-medoids clustering~\cite{Dinata2021}, Hierarchical agglomerative clustering~\cite{Wang2021}, non-negative matrix factorization (NMF) techniques were adopted along the LDA topic model to compare the clustering methods. The raw data from the three databases were preprocessed to remove hashtags, punctuations, and stop-words. The word embedding models weighted Word2vec, unweighted Word2vec, and Doc2vec were applied to all three datasets along with k- means clustering. The optimal number of epochs in each approach with its results were compared. All the methods were evaluated using three performance metrics normalized mutual information (NMI)~\cite{Rahmanian2022}, adjusted mutual information (AMI)~\cite{Hu2021}, and adjusted rand index (ARI) measures~\cite{DAmbrosio2020}. With their results, word embedding models outperformed traditional TF-IDF representations. The research work summarized end-to-end NLP tasks starting from data extraction to methods evaluation including data preparation, document clustering, and topic modelling. 

In a study conducted by Patil et al.~\cite{Patil2018}, aspect-level sentiment analysis was proposed to analyze e-commerce site Amazon product reviews~\cite{8597286,wassan2021amazon}. The authors extracted sentiment ratings from the website and categorized them into negative, neutral, and positive based on each product rating. In preprocessing step, tokenization and stemming techniques~\cite{Mystakidis2021} were implemented on the product reviews or user comments. LDA topic modelling technique and k-means clustering algorithm were used for topic extraction. Three machine learning models logistic regression~\cite{Palacios2021}, SVM~\cite{Gil2020}, and Naïve Bayes~\cite{GalopoPerez2021} were developed, one for each sentiment polarity. That model accuracy was calculated to know how the sentiment polarity worked for the textual data. Nine topics from electronic products reviews were extracted using an LDA and k-means clustering. Although that article focused on e-commerce product reviews, the process could be adopted for higher education feedback, where scores can be used to extract sentiment polarity and students comments to extract topics. Similarly, Kastrati et al.~\cite{9110884} proposed weakly supervised framework for aspect-level sentiment analysis and automatically identify sentiment or opinion in MOOC dataset. MOOC related aspects like content, structure, knowledge, skill, experience, assessment, technology, interaction, and general were grouped together to have four aspects for the proposed study. The four aspects are the course that covers both content and structure aspect, the instructor that includes knowledge, skill and experience of the instructor, the assessment, and the technology. Manually annotated dataset collected from Coursera was preprocessed with feature extraction technique TF-IDF, Word2Vec and CNN model was used for the aspect category learning. CNN and LSTM models were used for the aspect polarity assessment task with F1 score of 86.13\% for aspect category identification (broader MOOC-related aspects) and 82.10\% for aspect sentiment classification. 

A student feedback mining system (SFMS) for analyzing students' feedback was proposed by Gottipati et al.~\cite{Gottipati2018} based on agglomerative clustering with cosine similarity. Text analytics model was employed for text evaluation tasks like text categorization, entity extraction, sentiment analysis, document summarization. Ten topics were extracted using agglomerative clustering, and the top 5 to 10 words in each topic were used to label the topic. Logistic regression model was used for sentiment classification with a precision of 80.1\%, recall of 86.4\% and F-Score of 83.5\%. A faculty rating system based on text mining techniques was developed by Krishnaveni et al.~\cite{8126079}. Student feedback was mapped with student database and weights were assigned to student attributes like CGPA, sincerity, attendance, performance, and feedback submit duration. Naive Bayes classifier was used to rate faculty into classes range (1 star—5 star). 



Overall, LDA techniques~\cite{hamzah2020discovering,onah2021mooc,marccal2020strategy,Unankard2020,cunningham2019visualizing,perez2018international,srinivas2019topic,Curiskis2020,8681445,Ruan2022,8597286,wassan2021amazon,Patil2018} were the most used topic modelling methodology due to their generative process with the disambiguation of words in each topic and precise alignment of keywords to topics that may closely reflect the original collection.

\subsection{Text Evaluation}
In this subsection, NLP applications like text summarization, document categorization, text annotation, and knowledge graphs are discussed. 
\subsubsection{Text Summarization}
There has been an exponential growth in collection of student feedback for evaluation in educational institutions. Consolidating the content and extracting useful resources is a tedious task and would consume massive efforts. The summary of the content would be easier for readers to digest and comprehend. Text summarization technique provides a summary of a student feedback or corpus of the feedback text without losing critical information. Text summarization can be categorized into extractive, abstractive, and hybrid approaches~\cite{ElKassas2021}.

\begin{figure}[ht]
    \centering
    \includegraphics[width=\columnwidth, scale=2.0]{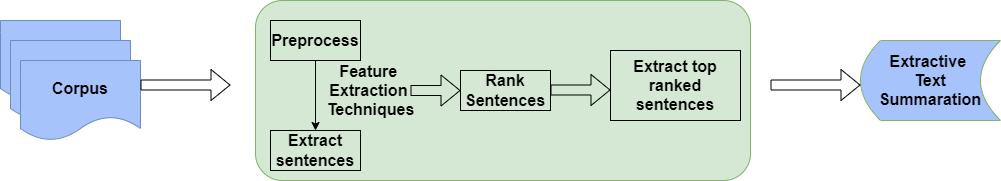}
    \caption{Extractive Text Summarization}
    \label{fig:ETS}
\end{figure}

\textbf{Extractive Text Summarization} is a traditional text summarization method. It extracts significant sentences as it is from the document and adds to the summary. As shown in Figure~\ref{fig:ETS}, the technique selects a subset of the sentences in an original text using feature extraction techniques like BoW, N-gram, graphs and so on. The extracted sentences are ranked based on their importance. It creates an intermediate representation that highlights the most important information included in the original text~\cite{allahyari2017text}.

\begin{figure}[ht]
    \centering
    \includegraphics[width=\columnwidth]{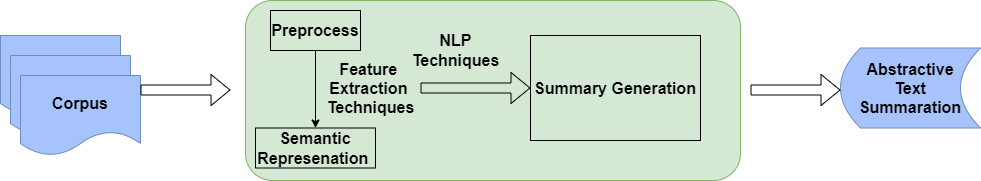}
    \caption{Abstractive Text Summarization}
    \label{fig:ATS}
\end{figure}

\textbf{Abstractive Text Summarization} extracts sentences from documents in an intermediate representation and generates a summary of the sentences instead of the original sentences as shown in Figure~\ref{fig:ATS}. The technique paraphrases the sentences using NLP techniques and generates a summary that is suitable to human interpretation~\cite{Abualigah2019}. 

\begin{figure}[ht]
    \centering
    \includegraphics[width=\columnwidth]{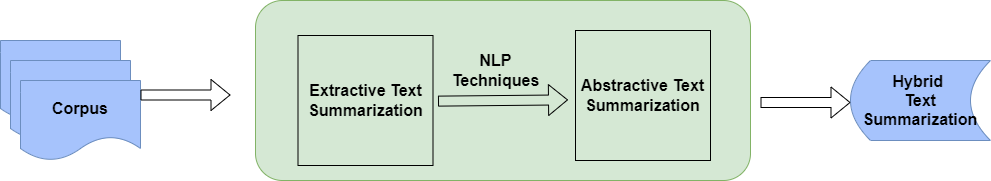}
    \caption{Hybrid Text Summarization}
    \label{fig:HTS}
\end{figure}

\textbf{Hybrid Text Summarization} is an ensemble of extractive and abstractive text summarization as shown in Figure~\ref{fig:HTS}. In this mechanism, the top-ranked sentences extracted from extractive text summarization are paraphrased using NLP techniques and summarizes the content abstractedly.

Mutlu et al.~\cite{Mutlu2020} stated that extractive sentiment has an advantage of language independence as it doesnot aim for sentence construction and paraphrasing. As stated earlier, it has intermediate representations where sentence scoring and sentence selection steps were involved. Each sentence is assigned a salience degree and ranked to summarize the important sentences in the original text. Estimating the salience of a sentence in the original text is a classification problem. The authors used a LSTM-NN (neural network) for the sentence selection based on semantic features, synthetic features and ensembled features. Compared the LSTM-NN model with the baseline models of hierarchical attention-based bidirectional gated recurrent unit (Bi-GRU), CNN, Bi-GRU and the newer state-of-the-art models SummaRuNNer~\cite{sefid2021extractive}, BanditSum~\cite{9364501}. The LSTM-NN model outperformed all other four models. Yuxiang et al.~\cite{wu2018learning} proposed a reinforced neural extractive text summarization model which optimizes the coherence and importance of summarized information simultaneously. The authors used a CNN model at the word level to extract features and their context. At the sentence level, a Bi-GRU was used to model the context of a sentence. The pre-trained data were fed to a reinforcement learning to compute cross sentence coherence as part of the reward of the proposed reinforced model. The research results showed that the proposed approach could balance cross-sentence coherence and sentence importance.

Statistical methods can be used for the extractive summarization process. Madhuri et al.~\cite{Madhuri2019} proposed a novel statistical method for extractive summarization. In the study, sentences were tokenized, stopwords removed, parts-of-speech added to each token, and the weights were assigned to each token based on its frequency and the total number of terms in the document. The weighted frequency of the token was calculated, and finally, the sum of the weighted frequency tokens was calculated. The sentences were rearranged in descending order and a summarizer extracts the highest rank sentences and converted them into an audio format. The work was evaluated with human summarized data and the proposed method achieved a higher accuracy. 

Fan et al.~\cite{Fan2015} proposed an CourseMIRROR (Mobile In-situ Reflections and Review with Optimized Rubrics) using automatic text summarization techniques to aggregate students' feedback. This approach assisted to extract most significant ones and help students to understand both difficulties and misunderstandings. The authors extracted phrases, grouped them using k-medoids clustering algorithm, and then re-rank the phrases by student coverage. This approach has generated better results than existing technique, LexRank~\cite{erkan2004lexrank}.

Gottipati et al.~\cite{gottipati2019topicsummary} proposed a topic based summarization tool to analyze student online discussion forum in a course and extract topic based summaries. The authors used TextRank Summarizer~\cite{mihalcea2004textrank} and LSA Summarizer~\cite{foong2015text} techniques for text summarization. Three questions were defined to categorize the online forum posts and extracts topics on each of these questions. LSA summary provided additional data recommendations when compared to TextRank summary. Luca et al.~\cite{8642858} proposed a methodology to recommend summaries of large teaching documents and these recommendations are customized to student's needs according to the test results conducted at end of lectures. A multiple-choice test was conducted at the end of a lecture to assess the student's level of understanding of different topics. The authors processed the text results and teaching material in parallel and summarized the content with multilingual weighted itemset-based summarizer (MWISum)~\cite{Baralis2015}. Based on the student understanding, teaching material summary was recommended. Similarly, a lecture summarization service was proposed by Miller~\cite{miller2019leveraging} using BERT model for dynamically sized lecture summarizations. The author used BERT model to generate embeddings for K-means clustering, which is an extractive text summarization approach. 

Abstractive text summarization preserves actual information and overall meaning while summarizing the sentences from a corpus with a shorter representation. In a study conducted by Song et al.~\cite{song2019abstractive}, a deep learning-based framework was proposed to construct new sentences based on semantic phrases using an LSTM-CNN model. The semantic phrases were not conventional tokenized sentences, the authors performed a phrase acquisition, phase refinement and phrase combination on the preprocessed database for the phrase extraction. The LSTM-CNN deep learning algorithm was trained using the extracted semantic phrase and set a threshold value to divide the text generation stages into a generating mode and a copy mode. The proposed approach of abstractive text summarization using the LSTM- CNN model outperformed other state-of-the-art systems using a CNN in terms of metric recall-oriented understudy for gisting evaluation (ROUGE-1)~\cite{royan2022text} of 34.9\% (an increase of 4.4\% over existing models) and the ROUGE-2~\cite{royan2022text} of 17.8\% (an increase of 1.6\% over existing models).

Asmussen et al.~\cite{Asmussen2019} developed a smart exploratory literature review where the authors proposed a three-step framework with pre-processing, topic modelling using an LDA technique, and post-processing. In the preprocessing step, articles were loaded to clean the non-value-adding words, to convert words to lowercases, to remove punctuations, special characters, whitespaces, URLs, and emails. The cleaning process differed from domain to domain, as non-value-adding words differed for each domain. Further to this, the number of topics in LDA topic modelling was estimated using a cross-validation technique. Once the number of topics was determined, the LDA model was executed. The outcomes of the model include the list of articles, a list of probabilities for each article for each topic, and a list of the most frequent words for each topic. In the post-processing step, identified research topics and labelled the topics that are relevant for use in a literature review. The LDA model was evaluated using statistical, semantic, or predictive approaches.

\subsubsection{Document Categorization}
Document categorization is one form of annotation to annotate a document in a text corpus~\cite{Boley1999}. It analyzes the content, intent and sentiment within a document and classifies them into predefined labels. Document categorization or text classification analogizes end-to-end entity linking where an entity linking labels individual words or phrases, document categorization annotates an entire text or body of a document with a single label. Sentiment annotation and linguistic annotation are part of document categorization to extract latent semantic and linguistic elements in a document.

Sindhu et al.~\cite{Sindhu2019} proposed supervised aspect-based opinion mining of students' feedback for teaching performance evaluation. In this study, six different aspects like teaching pedagogy, behaviour, knowledge, assessment, experience, and general were considered as domain understanding. Student feedback labelled with these aspects and description of each aspect were preprocessed to create academic domain word embeddings to represent words semantically. LSTM model with layer 1 as aspect extraction and layer 2 as opinion orientation was designed, and the model achieved accuracy of 91\% accuracy in aspect extraction and 93\% in sentiment detection.

Li et al.~\cite{Li2018} proposed an integrated hybrid deep learning methodology~\cite{Minaee2021,9451752} with a combination of LSTM and CNN models for Chinese text classification~\cite{Jang2020}. The features from processing serialized information in the LSTM were used along with a convolutional layer to extract more features. A BLSTM-C model was also proposed. The authors used three benchmark datasets in Chinese language with eight categories of articles. A BBC English news dataset with five categories was also tested to compare the experiment with Chinese datasets. All the datasets were preprocessed with word vectors using a Word2vec model and used a maxlen method to denote the maximum length of a sentence. Sentences with shorter lengths were padded with '0' vectors. The authors compared the classification accuracy of a simple LSTM and the proposed BLSTM-C on both English and Chinese datasets. They achieved an accuracy of 91.73\%, 94.88\% and 91.11\%, 96.23\% respectively.

\subsubsection{Entity Extraction}
To identify named entities, parts of speech and key phrases within a text, an entity annotation technique can be used~\cite{Cornolti2013}. Annotators read the text thoroughly to locate the target entities based on predefined labels. The located entities in entity annotation can be connected to larger repositories of data using entity linking. In end-to-end entity linking, preprocess a piece of text for named entity extraction. In entity disambiguation, extracted named entities will be linked to knowledge databases.

Dess et al.~\cite{Dess2021} proposed a novel architecture for extracting entities and relations among entities. An existing extractor framework~\cite{https://doi.org/10.48550/arxiv.1808.09602} based on a deep learning model and entity detection module was modified and embedded in the proposed architecture to detect six types of entities like task, method, material, metric, other scientific-term, and generic. Seven types of relations like compare, part-of, conjunction, evaluate-for, feature-f, hyponym-of, used-for were defined.

\subsubsection{Knowledge Graphs}
Knowledge graphs can represent information extracted using NLP in an abstract form and integrate the information extracted from multiple data sources. Domain knowledge from knowledge graphs are input into a machine learning model to produce better predictions. A knowledge graph can be served as a data structure which can store information. A combination of human input, automated and semi-automated extracted data can be added to a knowledge graph. 

\begin{figure}[!htp]
    \centering
    \includegraphics[width=\columnwidth]{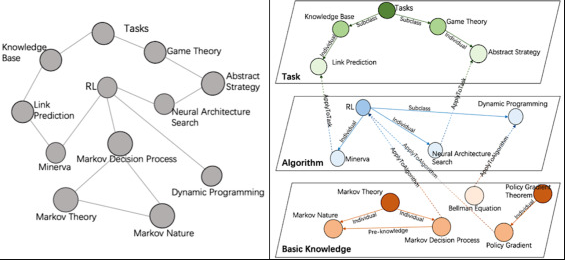}
    \caption{Sample Knowledge Graph~\cite{Shi2020}}
    \label{fig:kg}
\end{figure}

To recommend a well-organized diverse learning path, Shi et al.~\cite{Shi2020} proposed a learning path recommendation model based on a multidimensional knowledge graph framework. In this framework, learning objects were separately stored in several classes and it also proposed six semantic relationships between the learning objects in the knowledge graph. Figure~\ref{fig:kg} presents the multidimensional knowledge graph from~\cite{Shi2020} where dotted lines represent inter-class relationships, solid lines represent intra-class relationships, and the nodes with different colors represent learning objects in different classes. Then a learning path recommendation model was designed to traverse each from the knowledge graph and recommend the best learning path to students. Extracting context data in NLP is critical as it highly influences the model classification~\cite{Drpinghaus2019}. To store the extracted data, knowledge graph is the best approach and can easily map with different objects.

\subsubsection{Sentiment annotation}
One of the most trending annotations in NLP is sentiment annotation which is to label emotion, opinion and sentiment inherent within a text. The label could be a positive sentiment, neutral sentiment, or negative sentiment. It deals with emotional intelligence quotient in sentiment analysis or opinion mining. In natural language, understanding the context is critical. Without comprehensive understanding, it is difficult to predict the true emotion behind a text message or email. It is much more difficult for machines to mine customer intention in reviews or feedback, especially with sarcasm and humour. Sentiment annotated data are used to train machine learning models and help them to do sentiment analysis or opinion mining~\cite{Kastrati2021}\cite{https://doi.org/10.26220/une.2987}. 

Ibrahim et al.~\cite{ibrahim2018data} proposed a data mining framework to analyze student feedback with classification algorithms like Naive Bayes,
SVM, decision tree, and random forest. Sutoyo et al.~\cite{sutoyo2020sentiment} proposed a feedback questionnaire for lecturer evaluation based on student feedback to the questionnaire. Sentiment analysis was performed on the student feedback and classified them into positive or negative sentiment using CNN model. The deep learning model achieved accuracy, precision, recall, and F1-Score of 87.95\%, 87\%, 78\%, and 81\%, respectively. Similarly, Kandhro et al.~\cite{kandhro2019sentiment} proposed LSTM model with predefined word embedding layer for sentiment analysis and achieved an accuracy of 92\% and 79\% for positive and negative sentiment classification respectively. Using annotation technique, FIT-EBot, a chatbot in a university was proposed in a study conducted by Hien et al.~\cite{hien2018intelligent}. The authors used a NLP technique to extract context and intention in a student query for the chatbot. After analyzing the student's intention and context analysis, three models, namely a pattern-based model~\cite{joshi2015sarcasmbot}, a retrieval-based model~\cite{wu2016sequential}, and a generative model~\cite{patidar2017correcting} were used to build responses to the student query. While decoding student query, the text messages were classified into 13 predefined topics. A classifier was trained with the manually defined 13 topics based on the results from a survey. The 13 topics defined in the study was course registration, alternative course, prerequisite course, course content, major, course material, scholarship, graduation and others that were used to extract the intent of the student query~\cite{Lee2021}. After intention, the context of the query was extracted using named entity recognition. For that purpose, a corpus in which each word has been exactly identified with a label was trained using a classifier, so that the model could be used to extract context from the student query. The proposed approach achieved an F1-score of 82.33\%, 97.33\% for student intent identification and student context extraction respectively. A text annotation was executed to review students’ opinion and extracted the sentiment of the opinion in higher education in a study~\cite{grljevic2020opinion}. A MATTER (Model, Annotate, Test, Train, Evaluate, Revise) methodology was implemented as part of the annotation procedure. In that study, student opinion was annotated manually based on a predefined annotation scheme and evaluated using an inter-rater agreement. 


\section{Challenges}\label{challenges}

In this section, challenges in implementing NLP techniques in the education domain are discussed. 


\subsection{Domain-Specific Language}
In order to classify academic dataset or students' feedback, it is required to understand core factors of teaching context~\cite{Sindhu2019}. This is considered one of the challenges in implementing NLP in education domain. Considering abundant student feedback being generated from different surveys, questionnaires, and other educational feedback acquiring portals on a course teaching or a learning management system. Without understanding or getting trained on the specific domains, NLP methodologies could not be able to uncover the latent semantic meaning of a text. Nhi et al.~\cite{Vo2022} proposed a domain-specific NLP for students, faculty members, universities in computer science or information technology in higher education sector. The authors extracted tech-related skills using named entity recognition (NER) and built a personalized multi-level course recommendation system. This is a domain-specific NER designed to scrape data like job postings, course descriptions, and MOOCs online courses~\cite{Li2022} information from multiple websites and enhance the system with annotated corpus from StackOverflow and GitHub~\cite{tabassum2020code}. The annotated StackOverflow data were embedded and split into train, test and validation datasets. The train and test data were fed to a Bi-LSTM and the proposed CSIT-NER models for training, and GitHub data with StackOverflow test dataset was used to evaluate the models. The scraped data were embedded and extracted entities to form a corpus. Pashev et al.~\cite{Pashev2021} proposed a methodology to extract entities and their relations using MeaningCloud API and Google Translate API. The authors calculated grades based on the relevance to the topics created by a teacher or auto-generated text from the subject area. Extracting entities or concepts from a huge database available using a data scraping technique and processing them with considerable manual annotation would assist in building corpora for an application domain~\cite{Abulaish2018}.

\subsection{Sarcasm}
Decoding sarcasm is critical in NLP tasks like sentiment annotation and opinion analysis. This helps to decipher student opinions and perceptions on course structure and educational infrastructure. In a survey article in~\cite{joshi2017automatic}, automatic sarcasm detection was studied explicitly. The authors surveyed existing traditional sarcasm detection studies and reported the research gap. Sarcasm labels are hidden attributes that need to be predicted by considering conversations and sentences before and after a sarcastic text. The datasets used for sarcasm detection in that research were divided into categories like short text, long text, transcripts, dialogues and miscellaneous. To detect sarcasm, the authors reported three different approaches such as rule-based, statistical, and deep learning approaches. In a rule-based approach, sarcasm can be identified based on key indicators of sarcasm captured as evidence~\cite{Abulaish2018,Sundararajan2020, gaanoun-benelallam-2021-sarcasm}. In a statistical approach to detect sarcasm, punctuations, sentiment-lexicon-based features, unigrams, word embedding similarity, frequency of the rarest words, sentiment flips and so on were key features to the statistical classifiers~\cite{9120917,khatri-p-2020-sarcasm,8488096}. Traditional machine learning algorithms like SVM~\cite{garg2020sarcasm}, logistic regression, decision trees, Naive Bayes, hidden Markov model, and ensemble classification methods were also used in classifying the sarcasm. In deep learning algorithms, RNN models and LSTM methods~\cite{8949523} can be used individually as well as in combination with CNNs~\cite{10.1145/3277104.3277118} for automatic sarcasm detection. The survey article provided a comprehensive understanding of sarcasm detection.

\subsection{Ambiguity}
Ambiguity in natural languages is common as it depends on context and user perception in reading a text. With challenges in decoding a context, ambiguity in machine learning language processing is more complicated. Ambiguity could be due to the structure, syntactic, or lexical nature of a sentence~\cite{jusoh2018study}. In structural ambiguity, a sentence has more than one syntactic structure. In syntactic ambiguity, a grammatical construct error occurs in a sub-part of a sentence that causes grammatical ambiguity in a complete sentence. Lexical ambiguity is due to a word having two different meanings and two words having the same form. Addressing the ambiguity challenge is crucial in analyzing feedback. In a study in~\cite{yap2020adapting}, word sense disambiguation was addressed by customizing BERT, a language representation model, and selecting the best context-gloss pairs from a group of related pairs~\cite{huang-etal-2019-glossbert}. The authors classified the context-gloss pairs into positive and negative sentiment, and example sentences from WordNet 3.0 were combined with the positive and negative gloss pairs. Annotating the combination assisted in creating additional training samples. The proposed BERT model outperformed other existing state-of-art models in terms of F1-score with 77\%.

\subsection{Emoticons and Special Characters}
Emoticons and special characters play a vital role in opinion mining especially students’ feedback containing the special symbols to express their emotions. NLP has a challenging phase in processing the emoticons and labelling them with appropriate emotion tags. In a study in 2020~\cite{9207881}, the authors analyzed cross-cultural reactions to the novel coronavirus and detected sentiment polarity and emotion from their tweets and validated them with emoticons. A deep learning model based on LSTM was used in combination with feature extraction methods like GloVe, word embeddings. Six emotions of joy, surprise, sadness, anger, fear, and disgust were validated using different emoticons with their unicodes. Cappallo et al.~\cite{8424082} proposed a large dataset with real-world emojis and explained three challenges in emoticons processing. They were emoji processing, emoji anticipation, and query-by-emoji. The authors used two deep learning models, a Bi-LSTM model for text-to-emoji baseline results and a CNN model for image-to- emoji. They then combined the two algorithms to form a multi-modal method approach for emoticons processing. The work can be adopted in analyzing student opinions and processing the emoticons used in their feedback.

\subsection{Aspect-based Sentiment Analysis}
Chauhan et al.~\cite{Chauhan2018} quoted that sentiment analysis tool largely underused in education could not find opinions on different aspects. Most of the research works to process student comments or feedback to classify the positive or negative sentiment using lexicon-based or machine learning methods at document level. Nazir et al.~\cite{Nazir2020} conducted a survey on issues and challenges that are related to extraction of different aspects. The study was divided into three topics aspect extraction, aspect sentiment analysis, and sentiment evolution. Each topic was breakdown into sub-categories explicit aspect extraction, implicit aspect extraction, aspect level sentiment analysis, entity level sentiment analysis, multi-word sentiment analysis, recognition of factors in sentiment evolution, and predicting sentiment evolution over social data. 

\subsection{Data Imbalance}
Data imbalance is one of the most common challenges in AI~\cite{Thabtah2020} in which number of samples in one class exceeds the amount in other classes. Considering NLP, a subset of AI, the challenge is inherited. Especially in education domain, it is difficult to acquire of massive labelled data as it requires manual annotation from domain experts. Although the acquired labelled data fed to deep learning algorithms, the classification performance is biased due to data distribution discrepancy~\cite{8511076}. A potential tool to overcome this challenge could transfer learning~\cite{ruder-etal-2019-transfer}, where a deep learning model trained on a large corpus of student feedback to perform similar tasks on another data source. Other techniques could be sampling techniques~\cite{Shaikh2021} to under-sample majority classes or over-sample minority classes, which might demand text augmentation tasks~\cite{Shorten2021}.

\section{Discussion}\label{Discussion}
\begin{figure*}[!ht]
    \centering
    \includegraphics[scale=0.5]{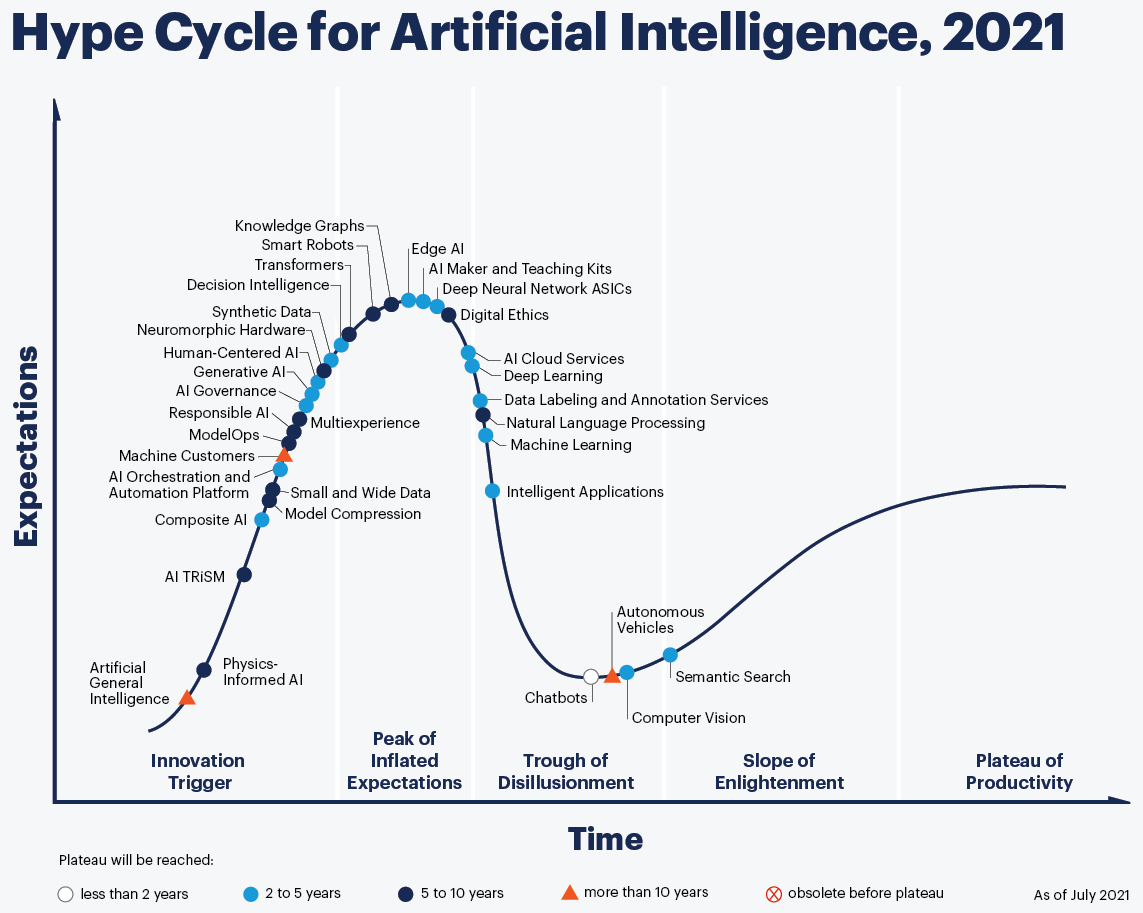}
    \caption{Gartner's Hype Cycle for Artificial Intelligence 2021~\cite{gartner}}
    \label{fig:Gartner}
\end{figure*}

According to a Gartner diagram shown in Figure~\ref{fig:Gartner}~\cite{gartner}, decision intelligence, deep learning, knowledge graphs are at peak point which can be adapted to the education domain to build decision support systems. These areas analyze existing data and streamline the process of data storage. Deep learning methods can be used without much expertise in the application domain and build semantic networks to store interlinked entity data in a domain. It is expected that 70\% of the organizations will shift their focus from big to small and wide data by 2025 to provide more context for data analytics~\cite{Almazmomi2021}. This infers a variety of small structured and unstructured data sources in diverse platforms.

%
\begin{table*}[!htp]
\caption{NLP - Programming packages}
\label{tab:NLP-PP}
\begin{tabular}{@{}cll@{}}
\toprule
\textbf{Programming Languages} &
  \multicolumn{1}{c}{\textbf{Packages}} &
  \multicolumn{1}{c}{\textbf{Features}} \\ \midrule
\multirow{9}{*}{\textbf{Python}} &
  spaCy &
  \begin{tabular}[c]{@{}l@{}}Part-of-Speech Tagging, Tokenization, Dependency Parsing, Sentence Segmentation, \\ Entity \& Sentence Recognition, Seamless integration with Deep Learning, \\ Methods for cleaning and normalizing text\end{tabular} \\ \cmidrule(l){2-3} 
 &
  CoreNLP&
  \begin{tabular}[c]{@{}l@{}}Lemmatization, Named Entity Recognition, Morphological Tagging, Tokenization, \\ Part-of-Speech tagging\end{tabular} \\ \cmidrule(l){2-3} 
 &
  NLTK &
  Tokenization, Part-of-Speech Tagging, Stemming \\ \cmidrule(l){2-3} 
 &
  Gensim &
  Topic Modelling Algorithms \\ \cmidrule(l){2-3} 
 &
  PyNLPI &
  \begin{tabular}[c]{@{}l@{}}Tokenization, Frequency Lists, Supports formats like FoLiA, \\ timbl, GIZA++, moses\end{tabular} \\ \cmidrule(l){2-3} 
 &
  Pattern &
  \begin{tabular}[c]{@{}l@{}}Part-of-Speech Taggers, n-grams, Sentiment Analysis, WordNet, \\ Text Classification, Tokenization\end{tabular} \\ \cmidrule(l){2-3} 
 &
  Polyglot &
  Tokenization, Language detection, part-of-speech tagging \\ \cmidrule(l){2-3} 
 &
  TextBlob &
  \begin{tabular}[c]{@{}l@{}}Part-of-Speech Tagging, Sentiment Analysis, Classification, Tokenization, n-grams, \\ Word Inflection, WordNet Integration, Language translation and detection powered \\ by Google Translate, Word, and phrase frequencies, Parsing, Spelling correction, \\ Add new models or languages through extensions\end{tabular} \\ \cmidrule(l){2-3} 
 &
  Quepy &
  Transform natural language questions to queries in a database query language \\ \midrule
\multirow{6}{*}{\textbf{Java}} &
  OpenNLP &
  \begin{tabular}[c]{@{}l@{}}Lemmatization, Named Entity Recognition, Morphological Tagging, Tokenization, \\ Part-of-Speech tagging\end{tabular} \\ \cmidrule(l){2-3} 
 &
  StanfordNLP &
  \begin{tabular}[c]{@{}l@{}}Lemmatization, Named Entity Recognition, Morphological Tagging, Tokenization, \\ Part-of-Speech tagging\end{tabular} \\ \cmidrule(l){2-3} 
 &
  CogCompNLP &
  \begin{tabular}[c]{@{}l@{}}Collects a number of core libraries for NLP for Part-of-Speech Tagging, \\ Lemmatization, Named Entity Recognition, Temporal extractor and normalizer\end{tabular} \\ \cmidrule(l){2-3} 
 &
  MALLET &
  \begin{tabular}[c]{@{}l@{}}Statistical NLP with document classification, clustering, topic modelling, \\ information extraction\end{tabular} \\ \cmidrule(l){2-3} 
 &
  NLP4J &
  \begin{tabular}[c]{@{}l@{}}Morphological analysis, Data Crawler, Annotator, Parsing, \\ Statistical indexing function\end{tabular} \\ \cmidrule(l){2-3} 
 &
  CORENLP &
  \begin{tabular}[c]{@{}l@{}}Derive linguistic annotations for text, Part-of-Speech tagging, Named Entity \\ Recognition, Numeric and Time values, Dependency and Constituency parses, \\ Co-reference, Sentiment, Quote attributions, and relations\end{tabular} \\ \midrule
\multirow{5}{*}{\textbf{R Programming}} &
  koRpus &
  \begin{tabular}[c]{@{}l@{}}Part-of-Speech tagging, Tree Tagger, functions for automatic language detection, \\ hyphenation, several indices of lexical diversity\end{tabular} \\ \cmidrule(l){2-3} 
 &
  Isa &
  \begin{tabular}[c]{@{}l@{}}Latent Semantic Analysis, statistically derive conceptual indices \\  to extract higher order structure.\end{tabular} \\ \cmidrule(l){2-3} 
 &
  OpenNLP &
  \begin{tabular}[c]{@{}l@{}}Lemmatization, Named Entity Recognition, Morphological Tagging, \\ Tokenization, Part-of-Speech tagging\end{tabular} \\ \cmidrule(l){2-3} 
 &
  Quanteda&
  \begin{tabular}[c]{@{}l@{}}Quantitative analysis of textual data , tokenizers, \\ ngrams, and analyzing keywords\end{tabular} \\ \cmidrule(l){2-3} 
 &
  RWeka&
  \begin{tabular}[c]{@{}l@{}}Collection of machine learning algorithms. It deals with data mining, \\ data pre-processing,classification, regression, clustering, \\ association rules, and visualization\end{tabular} \\ \bottomrule
\end{tabular}
\end{table*}


NLP, part of AI, makes it possible to understand human language and listen to their opinions and feedback. Especially, education institutions need to adopt NLP methods to enhance the student learning experience, personalized learning management systems~\cite{eggert2021artificial}, and teacher training. This would help transform and get students expose to AI- driven tools. There is a wide variety of AI applications in education like smart classrooms with video \& audio data annotation~\cite{8253436,gerritsen2018towards, Steinbauer2021}, NLP for textual data annotation, classification, summarization, and image processing to detect gestures~\cite{Timms2016,chen2021building}. In this study, NLP methodologies were focused on and discussed their applications by AI. Although few research articles are related to AI in education directly, their approaches can be adapted to education.

Holmes et al.~\cite{holmes2019artificial} discussed problems and future implications of AI in education. The authors raised two problems of ”What we teach, and how we teach it”. What we teach refers to what students should learn in the age of AI, and the learning goals should be versatility, relevance, and transferability. To achieve these goals, strategies like emphasize on selective traditional knowledge areas, addition of modern knowledge, interdisciplinary concepts, embedded skills, and meta learning were proposed. Coming to the how-to question, it refers to how AI can enhance and transform education. The authors draw a line between education technology and AI in education. Education technology is to amend the taxonomy and ontology of the field, whereas AI in educations deals with a layered framework of substitution, augmentation, modification, and redefinition, which includes enhancement and transformation. The authors quoted an assumption where many assume robot teachers teach students with AI in education~\cite{Felix2020,selwyn2019should}. Although it could be possible in the future, current research is to transform and evolve the education industry to amplify the student learning experience without considering the enhancement of teaching practices. The authors proposed intelligent tutoring systems which involved a domain model with subject knowledge, a pedagogy model with effective teaching and learning approaches, a learner model for individual student learning. 

NLP techniques can be executed on feedback data using different programming languages. Even though there are different languages with pre-built packages to execute NLP methods, Python, Java and R programming languages are widely being used~\cite{Egger2022}. The factors to be considered in adopting a programming language includes the expertise in the specific programming language, the number of libraries or package tools that could assist in performing NLP tasks~\cite{Thomas2021}. Python~\cite{sarkar2019text}, with its versatility and simple consistent syntax mirroring human language, can offer a huge number of NLP packages for topic modelling, word embeddings, document classification, sentiment annotations. Java~\cite{reese2018natural} is a platform-independent language with robust architecture that can provide comprehensive text analysis tasks like clustering, tagging, and information extraction using multiple packages. R programming language~\cite{jockers2020text} is popular for its statistical learning, it is also being widely used in NLP tasks. The programming language can handle computationally intensive data analytics and investigate big data applications. Table~\ref{tab:NLP-PP} presents the three programming languages with their NLP packages. It has features for each package and its documentation source. 

In this work, four research questions on NLP in education were explored. For the first research question, the existing methodologies being used for NLP were discussed. Section~\ref{Methods} discussed in details data preprocessing methods feature extraction and feature selection definitions, types, and research community's work using machine learning and deep learning models. Different approaches of machine learning technique topic modelling were explained and student feedback topic extraction works were explored. Further to this, text evaluation techniques like text summarization, document categorization, text annotation, and knowledge graphs were explained.

The second research question was about the challenges of NLP in the education domain. Generic NLP challenges like domain-specific language, sarcasm, ambiguity, data imbalance are challenging in education to uncover the latent semantic meaning of students' feedback. The research community addressed this challenge using NER, rule-based, statistical, deep learning and BERT modelling. Emoticons and special characters are used to express their sentiment in feedback. To process these special symbols and characters, a multimodal approach was used. Converting emojis to their corresponding unicodes or image processing were used to determine the sentiment. Aspect-based sentiment analysis is more trending challenge of NLP in education domain. This challenge also refers to fine-grained sentiment analysis~\cite{Kastrati2021}.  

The third research question is to address the trends in NLP methodologies that could be adopted in the education domain. Those language processing methods were discussed in detail. AI models need to be trained in a quantitative approach, which will require pre-processing the textual data into vectors using feature extraction and feature selection techniques. In topic modelling techniques, both probabilistic and non-probabilistic models were discussed but the LDA technique from probabilistic models was the most commonly used to extract unsupervised topics from a corpus. Research on the fourth question investigated the community's work from other industry applications with short text analysis was discussed to understand~\cite{Deepa2019, Curiskis2020} and adapt to the education application domain. In \cite{Deepa2019}, twitter text analysis is short text analysis using VADER and SentiWordNet techniques. This approach can be applied to student feedback analysis, which would be short text in most cases. Query expansion ranking in \cite{parlar2018qer} is feature selection method that can be used in education feedback analysis. Patil et al.~\cite{Patil2018} approach to analyze e-commerce site Amazon product reviews can be directly used in higher education feedback analysis to analyze student ratings to a course delivery and extract sentiment aspects from student comments. Emojis are one of the forms' that student use to express their opinion in feedback. The approach in \cite{8424082} can be used to process emojis in student comments.


\begin{table*}[!htp]
\centering
\caption{NLP Techniques—Research Works}
\label{tab:References}
\begin{tabular}{@{}lll@{}}
\toprule
\textbf{NLP Techniques} & \textbf{Algorithms} & \textbf{Research Works} \\ \midrule
\multicolumn{2}{l}{\textbf{Methodology}} \\ \midrule
\multicolumn{1}{l}{Feature Extraction} & \multicolumn{1}{l}{BERT, LSA, LDA, CNN, LSTM, LSTM+ATT }& \multicolumn{1}{l}{\cite{Ahuja2019,waykole2018review,Dzisevic2019,Deepa2019,Goularas2019,Shah2020,goldberg69neural,Amin2020,Onan2020,Balamurali2020,ArroyoFernndez2019,8999624,Joseph2020,devlin2018bert,hou2020bert,zhang2020bert,masala2021extracting,wu2021word,Sangeetha2020}} \\ \midrule
\multicolumn{1}{l}{Feature Selection} & \multicolumn{1}{l}{Naive Bayes, Chi-square, SVM, random forest}& \multicolumn{1}{l}{\cite{ghojogh2019feature,cai2018,kou2020,wang2020,bommert2020,dhal2021,osman2017,parlar2018qer,Nikoli2020,8483345,7924899,gutierrez2018mining,9268731}} \\ \midrule
\multicolumn{1}{l}{Topic Modelling} & \multicolumn{1}{l}{LSA, NMF, LDA, PLSA}& \multicolumn{1}{l}{\begin{tabular}[c]{@{}l@{}}\cite{nikolenko2016,de2017theory,hamzah2020discovering,onah2021mooc,marccal2020strategy,perez2018international,cunningham2019visualizing,srinivas2019topic,Patil2018,Curiskis2020,Wallach2006,8250563,Landauer1998,Lee1999,chen1996building,Hofmann2001,Singh2020,Unankard2020,8681445,Ruan2022,8597286,wassan2021amazon,Liermann2021,Rani2021,Dinata2021,Wang2021,Rahmanian2022,Hu2021,DAmbrosio2020,Mystakidis2021,Palacios2021,Gil2020,GalopoPerez2021,8190703,9110884,8126079,wang2017deep,fowler2021autograding,zhang2021math,Gottipati2018,8658457,loria2018textblob }\end{tabular}} \\ \midrule
\multicolumn{2}{l}{\textbf{Text Evaluation}} \\ \midrule
\multicolumn{1}{l}{Text Summarization} & \multicolumn{1}{l}{LSTM-NN, CNN, Bi-GRU, TextRank, LSA, BERT, LSTM-CNN}& \multicolumn{1}{l}{\begin{tabular}[c]{@{}l@{}}\cite{ElKassas2021,allahyari2017text,Mutlu2020,Madhuri2019,song2019abstractive,wu2018learning,Asmussen2019,Abualigah2019,sefid2021extractive,9364501,royan2022text,Fan2015,erkan2004lexrank,gottipati2019topicsummary,mihalcea2004textrank,foong2015text,8642858,Baralis2015,miller2019leveraging}\end{tabular}} \\ \midrule
\multicolumn{1}{l}{\begin{tabular}[c]{@{}l@{}}Document\\ Categorization\end{tabular}} &\multicolumn{1}{l}{LSTM, CNN, BLSTM-C}& \multicolumn{1}{l}{\begin{tabular}[c]{@{}l@{}}\cite{Li2018,Boley1999,Jang2020,Minaee2021,9451752,Sindhu2019}\end{tabular}} \\ \midrule
\multicolumn{1}{l}{Entity Extraction} & \multicolumn{1}{l}{CNN}& \multicolumn{1}{l}{\cite{Cornolti2013,Dess2021,https://doi.org/10.48550/arxiv.1808.09602}} \\ \midrule
\multicolumn{1}{l}{Knowledge Graphs} &\multicolumn{1}{l}{Multidimensional knowledge graphs}& \multicolumn{1}{l}{\cite{Drpinghaus2019,Shi2020}} \\ \midrule
\multicolumn{1}{l}{\begin{tabular}[c]{@{}l@{}}Sentiment \\ Annotation\end{tabular}} & \multicolumn{1}{l}{CNN, LSTM}& \multicolumn{1}{l}{\cite{hien2018intelligent,joshi2015sarcasmbot,wu2016sequential,patidar2017correcting,grljevic2020opinion,Kastrati2021,Lee2021,https://doi.org/10.26220/une.2987,ibrahim2018data,sutoyo2020sentiment,kandhro2019sentiment}} \\
\bottomrule
\end{tabular}
\end{table*}

\begin{figure}
    \centering
    \includegraphics[width=\columnwidth]{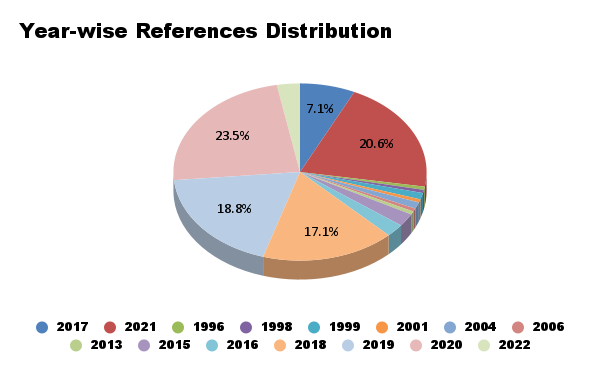}
    \caption{Year-wise References Distribution}
    \label{fig:Year-wise_References_Distribution}
\end{figure}

\begin{figure}
    \centering
    \includegraphics[width=\columnwidth]{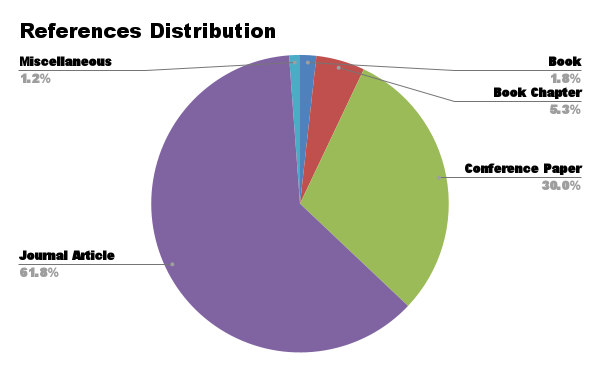}
    \caption{References Distribution}
    \label{fig:References_Distribution}
\end{figure}
\section{Conclusions} \label{Conclusions}
The aim of the study is to explore existing NLP methodologies that can be implemented or adopted in education domain. This assist to understand AI impact on education with open opportunities, synthesize the methods to process student feedback, and annotate their views. The literature review has been performed using Google Scholar covering bibliographic databases such as Wiley, Scopus, Springer, ACM Digital Library, IEEE Xplore, Pub-Med, Science Direct, and Multidisciplinary Digital Publishing Institute (MDPI) and so on. The search results of Google Scholar were manually checked for relevance NLP techniques in student feedback or education applications that can be adopted to the feedback analysis. For example, Twitter data analysis which consists of short text analysis using NLP similar student feedback. As shown in Figure~\ref{fig:Year-wise_References_Distribution}, the majority of the references included in this study are from the last 5 years. Also, more than 90\% percent of the citation included in this study are journal articles and conference papers. Table~\ref{tab:References} presents the NLP techniques explored in this study and corresponding research community works citations.

In this review article, the impact of AI on education was discussed. The scope of introducing AI into educational institutions is detailed based on the opportunities. Limiting the scope of introducing NLP methodologies to education for feedback analysis in this article, existing NLP methodologies were explored. Feature extraction, feature selection and topic modelling methodologies were explained with brief definitions. Further to this, text evaluation techniques text summarization, annotation, and knowledge graphs were reviewed. Each of these applications was defined and existing approaches were discussed. Challenges in adopting NLP methodologies to the education domain were reviewed. The limitation of this research is that this study is confined to AI implementation methodologies with less focus on pedagogy concepts. Data specific challenges like data scarcity and class imbalance were not discussed. This would affect the model learning for deep learning algorithms, which are data hungry. Strategies to interpret deep learning models (black box) were not explored. The future direction of this research would be to explore data challenges while extracting feedback or opinions without affecting privacy.

\textbf{Author Contributions:}
Conceptualization, T. Shaik. X. Tao. C. Dann; Methodology,  T. Shaik. X. Tao. and Y. Li; Formal Analysis, T. Shaik. X. Tao; Investigation, T. Shaik. X. Tao, Y. Li, C. Dann, P. Redmond, J. McDonald, L. Galligan; Data Curation, T. Shaik.; Writing – Original Draft Preparation, T. Shaik. X. Tao; Writing – Review \& Editing, T. Shaik. X. Tao, Y. Li, C. Dann, J. McDonald, P. Redmond, L. Galligan; Supervision, X. Tao, C. Dann; Project Administration, J. McDonald; Funding Acquisition, C. Dann 

\textbf{Conflicts of Interest:} The authors declare no conflict of interest. 

\bibliographystyle{IEEEtran}
\bibliography{biblio}

\begin{IEEEbiography}[{\includegraphics[width=1in,height=1.25in,clip,keepaspectratio]{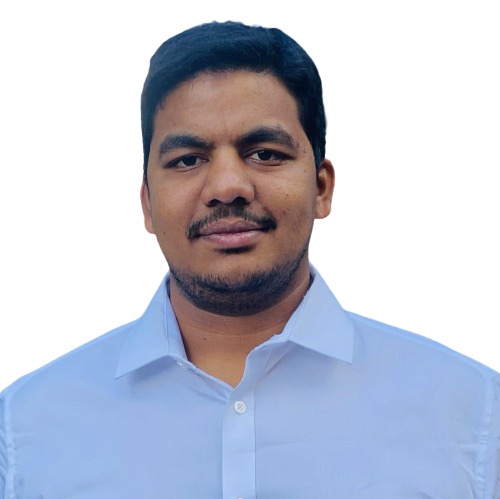}}]{Thanveer Shaik} is a PhD student at University of Southern Queensland, Australia. He finished his master’s degree with applied data science as major at USQ. His research interests are cognitive computing, biometrics, and NLP with expertise in Artificial Intelligence (AI), machine learning, and predictive analysis. 
\end{IEEEbiography}
\begin{IEEEbiography}[{\includegraphics[width=1in,height=1.25in,clip,keepaspectratio]{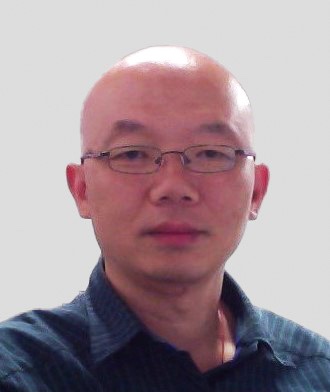}}]{Dr. Xiaohui Tao} is an elected Senior Member of IEEE and ACM, an active researcher in AI, and Associate Professor (Computing) in School of Mathematics, Physics and Computing (SoMPC), University of Southern Queensland (USQ), Australia. His research interests include data analytics, machine learning, knowledge engineering, information retrieval, and health informatics. His research outcomes have been published on many top-tier journals (e.g., TKDE, IPM, KBS, ESWA, and PRL) and conferences (e.g., IJCAI, ICDE, CIKM, PAKDD and WISE). Tao received ARC DP grant (Ref. DP220101360) in 2022-24, Australian Endeavour Research Fellow in 2015-16, and was awarded with ``Research Performance Award'' and ``Discipline Research Performance Improvement Award'' by SoMPC, USQ, among many others. Tao has been active in professional services. He has served PC Chair in WI '17, '18, WI-IAT '21 and BESC '18 and '21 and an editor or guest editor in many journals including INFFUS and WWWJ. Tao was awarded PhD in Queensland University of Technology, Brisbane.
\end{IEEEbiography}
\begin{IEEEbiography}[{\includegraphics[width=1in,height=1.25in,clip,keepaspectratio]{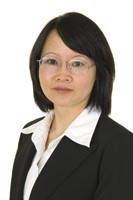}}]{Yan Li} is a Professor in Computer Science with the School of Mathematics, Physics and Computing at the University of Southern Queensland, Australia. Her research interests include artiﬁcial intelligence, big data analytics, signal and image processing, biomedical engineering, and computer networking technologies and security. 
\end{IEEEbiography}
\begin{IEEEbiography}[{\includegraphics[width=1in,height=1.25in,clip,keepaspectratio]{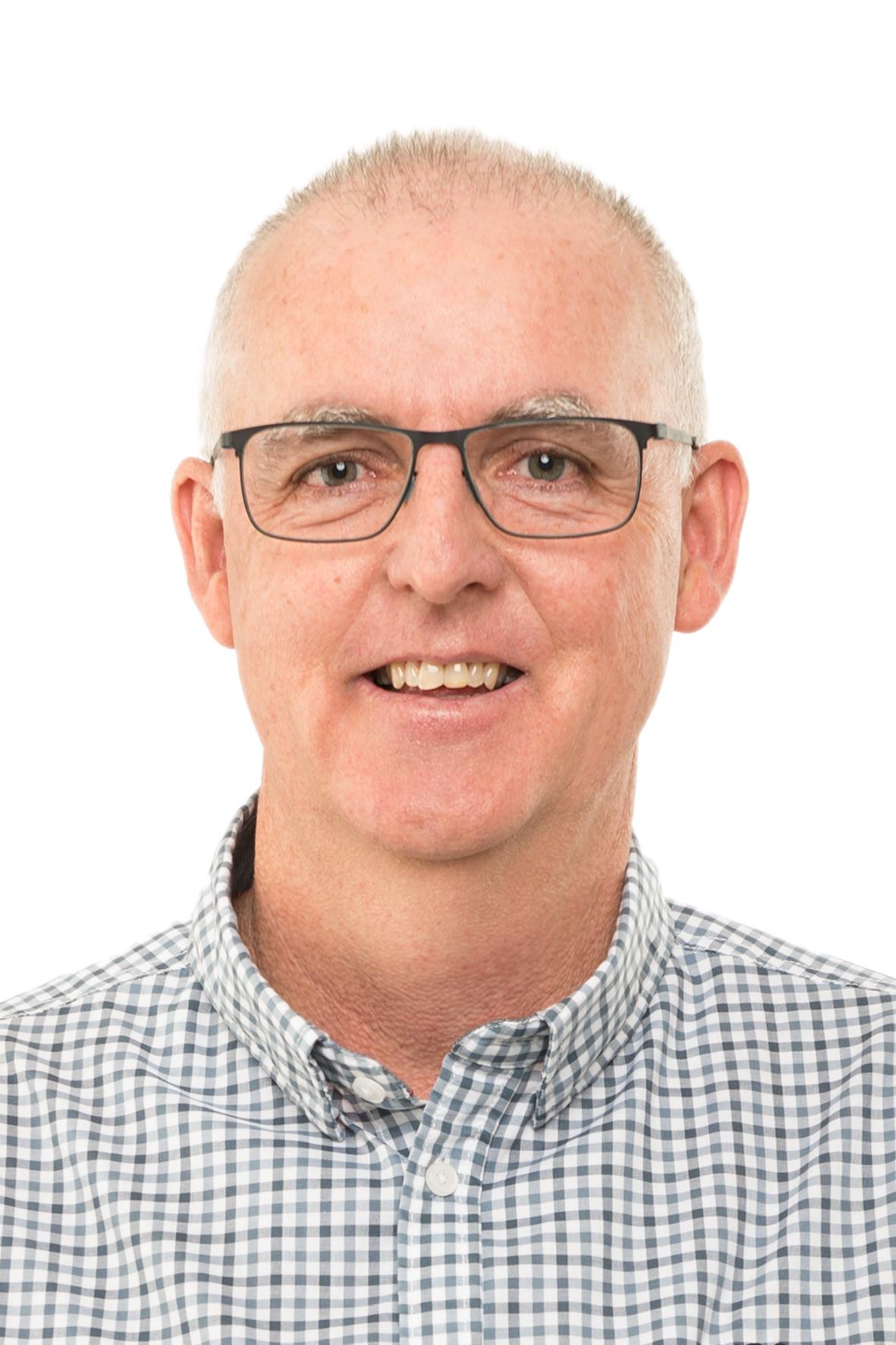}}]{Christopher Dann} is an inclusive goal orientated leader whose purpose is to make a positive impact on the educational experiences of learners ‘glocally’.  Chris is currently a Senior Lecturer at the University of Southern Queensland, Curriculum and Pedagogy (Technologies), in the School of Teacher Education. His current research is exploring the possible impact of Machine Learning and Artificial Intelligence on the teaching and learning process from the perspective of teachers and students across educational context. 
\end{IEEEbiography}
\begin{IEEEbiography}[{\includegraphics[width=1in,height=1.25in,clip,keepaspectratio]{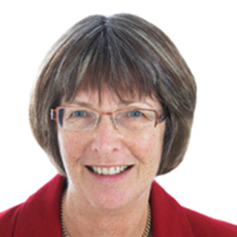}}]{Jacquie McDonald} is an Honorary Associate Professor at the University of Southern Queensland (USQ) Australia and a Higher Education Community of Practice (CoP) consultant. She previously worked for over 26 years as a Learning and Teaching Designer at USQ designing online and distance learning courses and programs. Since 2006 she has facilitated, researched and coached the implementation of inter/national Higher Education CoPs and led a number of institutional and national fellowships and grants. She is a member of the Australian Learning and Teaching Fellows Alumni and is on the editorial board of the Journal on Excellence in College Teaching. Jacquie has won university and national awards, citations and best paper awards at international conferences. Her research and publications focus on social learning spaces, including Communities of Practice, designing for online learning and teaching, and educational professional development. Recent publications include Springer (2017) co-edited books, Communities of Practice: Facilitating Social Learning in Higher Education and Implementing Communities of Practice in Higher Education: Dreamers and Schemers, and (2021), Sustaining Communities of Practice with Early Career Teachers: Supporting Early Career Teachers in Australian and International Primary and Secondary Schools.  
\end{IEEEbiography}
\begin{IEEEbiography}[{\includegraphics[width=1in,height=1.25in,clip,keepaspectratio]{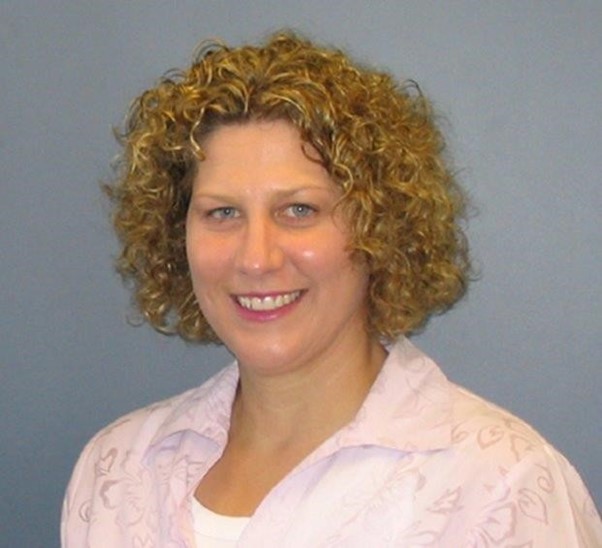}}]{Professor Petrea Redmond} is currently the Associate Head of School (Education), Research.  She completed her PhD (2011) in Educational Technology at the University of Southern Queensland.  Prior to that, she completed a Master of Education, a Bachelor of Business and a Diploma of Teaching. She has worked in high schools and in universities in Australia and Canada.  

Professor Redmond has over 80 publications, including two edited books, and has been the editor of the Australasian Journal of Educational Technology (AJET). Her current teaching and research interests include AI and education, telepresence robots, online engagement and digital pedagogies. She is currently a member of the executive for the Australasian Society for Computers in Learning in Tertiary Education (ASCILITE) and The Society for Information Technology and Teacher Education (SITE).  

Professor Redmond has received national and international awards for publications and research programs including best paper awards at international conferences: short listed for UKs Association for Learning Technology (ALT) Learning Technologist of the Year Team Award; USQ excellence award for Advancing Student Success; USQ excellence award for Online Learning Innovation, Queensland Government Our Women, Our State Award; and the Australasian Society for Computers in Learning in Tertiary Education (ASCILITE) Fellow Award.
\end{IEEEbiography}
\begin{IEEEbiography}[{\includegraphics[width=1in,height=1.25in,clip,keepaspectratio]{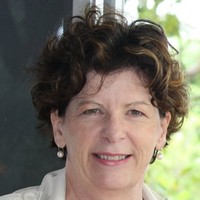}}]{Linda Galligan} is experienced Associate Professor with a demonstrated history of working in the higher education industry. Skilled, within the context of mathematics, in Lecturing, Educational Technology, Educational Research, and Applied Linguistics. Strong education professional with a PhD focused in Mathematics Education from QUT. 

\end{IEEEbiography}

\EOD

\end{document}